\documentclass[runningheads,a4paper]{llncs}
\usepackage{times}
\usepackage{graphicx}
\usepackage{amsmath,amssymb}
\usepackage{color}
\usepackage{url}
\usepackage{xspace}
\usepackage{algorithmic}
\usepackage{booktabs}
\usepackage{enumerate}
\usepackage{algorithm}
\usepackage{float}
\usepackage{rotating}
\usepackage{multirow}

\usepackage{hyperref}
\hypersetup{colorlinks=true,linkcolor=black,citecolor=blue,filecolor=black,urlcolor=blue}

\graphicspath{{./}}

\newcommand{\ie}{{ie.}\@\xspace}
\newcommand{\eg}{{eg.}\@\xspace}
\newcommand{\etal}{{et~al.}\@\xspace}

\begin{document}
\mainmatter
\def\ECCV12SubNumber{325}  

\title{Sparse Coding and Dictionary Learning for Symmetric Positive Definite Matrices: A Kernel Approach}

\titlerunning{Sparse Coding and Dictionary Learning: A Kernel Approach}

\authorrunning{Harandi et al.}

\author
  {
  Mehrtash~T.~Harandi, 
  Conrad~Sanderson,
  Richard~Hartley,
  Brian~C.~Lovell
  }

\institute
  {
  NICTA, PO Box 6020, St Lucia, QLD 4067, Australia%
  ~\thanks
    {
    Published in: {\bf Lecture Notes in Computer Science (LNCS), Vol.~7573, pp. 216--229, 2012.}
    \url{http://dx.doi.org/10.1007/978-3-642-33709-3_16}
    }%
  \\
  University of Queensland, School of ITEE, QLD 4072, Australia\\
  NICTA, Locked Bag 8001, Canberra, ACT 2601, Australia\\
  Australian National University, Canberra, ACT 0200, Australia
  }

\maketitle
\thispagestyle{empty}
\pagestyle{empty}

\begin{abstract}

Recent advances suggest that a wide range of computer vision problems
can be addressed more appropriately by considering \mbox{non-Euclidean} geometry.
This paper tackles the problem of sparse coding and dictionary learning
in the space of symmetric positive definite matrices,
which form a Riemannian manifold.
With the aid of the recently introduced Stein kernel
(related to a symmetric version of Bregman matrix divergence),
we propose to perform sparse coding by embedding Riemannian manifolds into reproducing kernel Hilbert spaces.
This leads to a convex and kernel version of the Lasso problem,
which can be solved efficiently.
We furthermore propose an algorithm for learning a Riemannian dictionary (used for sparse coding),
closely tied to the Stein kernel.
Experiments on several classification tasks
(face recognition, texture classification, person re-identification)
show that the proposed sparse coding approach
achieves notable improvements in discrimination accuracy,
in comparison to state-of-the-art methods such as
tensor sparse coding, Riemannian locality preserving projection,
and symmetry-driven accumulation of local features.
\end{abstract}

\section{Introduction}
\label{sec:introduction}

Sparse representation (SR),
the linear decomposition of a signal using a few atoms of a dictionary,
has led to notable results for various image processing and computer vision tasks~\cite{ELAD_SR_BOOK_2010,Wright_2009_PAMI}.
While significant steps have been taken towards expanding the theory of SR,
such representations in non-Euclidean spaces have received comparatively little attention.
This paper tackles the problem of sparse coding within the space of symmetric positive definite (SPD) matrices.

SPD matrices are fundamental building blocks in computer vision and machine learning.
A notable example is the covariance descriptor~\cite{Tuzel_2008_PAMI},
which offer a compact way of describing regions/cuboids in images/videos and fusion of multiple features.
Covariance descriptors have been exploited in several applications,
such as diffusion tensor imaging~\cite{Pennec_jmiv06},
action recognition~\cite{Lui2011,SR_Riemannian_AVSS_2010,Yuan:ACCV:2010},
pedestrian detection~\cite{Tuzel_2008_PAMI},
face recognition~\cite{PANG:TCSVT:2008,Sivalingam:ECCV:2010},
texture classification~\cite{Sivalingam:ECCV:2010,Harandi_WACV_2012},
and tracking~\cite{Hu:PAMI:2012}.

SPD matrices form a cone of zero curvature and can be analysed using the geometry of Euclidean space.
However, several studies have shown that a Riemannian structure of negative curvature is more suitable for analysing SPD matrices~\cite{Pennec_jmiv06,BHATIA_2007}.
More specifically, Pennec \etal~\cite{Pennec_jmiv06} introduced the Affine Invariant Riemannian Metric (AIRM)
and showed that the induced Riemannian structure is invariant to inversion and similarity transforms.
The AIRM is perhaps the most widely used similarity measure for SPD
matrices. Nevertheless, efficiently and accurately handling the Riemannian structure is non-trivial
as basic computations on Riemannian manifolds (such as similarities and distances) involve non-linear operators.
This not only hinders the development of optimisation algorithms but also incurs a significant numerical burden.

To address the above drawbacks,
in this paper we propose to perform the sparse coding of SPD matrices
by embedding Riemannian manifolds into reproducing kernel Hilbert spaces (RKHS)~\cite{Shawe-Taylor:2004:KMP}.
This is in contrast to directly embedding into Euclidean spaces~\cite{Yuan:ACCV:2010,SR_Riemannian_AVSS_2010,Sra:2011:ECML}.

{\bf Related Work.}
Sra \etal~\cite{Sra:2011:ECML} used the cone of SPD matrices
and the Frobenius norm as a measure of similarity between SPD matrices.
While this results in a regularised non-negative least-squares approach,
it does not consider the Riemannian geometry induced by AIRM.

Guo \etal~\cite{SR_Riemannian_AVSS_2010} and Yuan \etal~\cite{Yuan:ACCV:2010} separately proposed
to solve sparse representation by a log-Euclidean approach,
where a Riemannian problem is converted to an Euclidean one
by embedding manifolds into tangent spaces.
While log-Euclidean approaches benefit from simplicity,
the true geometry of the manifold is not taken into account.
More specifically, on a tangent space only distances to the pole of space are true geodesic distances.
As such, the pairwise distances between arbitrary points on the tangent space do not represent the structure of the manifold.

Sivalingam \etal~\cite{Sivalingam:ECCV:2010} used Burg divergence~\cite{Kulis:2009:JMLR} as a metric
and reformulated the Riemannian\footnote{We loosely use `Riemannian' to refer to the Riemannian manifold formed by SPD matrices.} 
SR problem as a determinant maximisation problem.
This has the advantage of avoiding the explicit manifold embedding,
as well as resulting in a convex MAXDET problem~\cite{Boyd:2004} that can be solved by interior point methods.
However, there are two downsides: the solution is computationally very expensive,
and the relations between Burg divergence and the geometry of Riemannian manifolds were not well established.

{\bf Contributions.}
With the aid of the recently introduced Stein kernel~\cite{Sra:JMLR:2012},
which is related to AIRM via a tight bound,
we propose a Riemannian sparse solver by embedding Riemannian manifolds into RKHS.
We show that the embedding leads to a convex and kernelised version of the Lasso problem~\cite{ELAD_SR_BOOK_2010},
which can be solved efficiently.
We furthermore propose a sparsity-maximising algorithm for dictionary learning within the space of SPD matrices,
closely tied to the Stein kernel.
Lastly, we show that the proposed sparse coding approach
obtains superior performance on several visual classification tasks
(face recognition, texture classification, person re-identification),
in comparison to several state-of-the-art methods:
tensor sparse coding~\cite{Sivalingam:ECCV:2010},
log-Euclidean sparse representation~\cite{SR_Riemannian_AVSS_2010,Yuan:ACCV:2010},
Gabor feature based sparse representation~\cite{GSR:ECCV2010},
and Riemannian locality preserving projection~\cite{Harandi_WACV_2012}.

We continue the paper as follows.
Section~\ref{sec:background} begins with an overview of Bregman divergence and the Stein kernel.
Section~\ref{sec:kernel_SR} describes the proposed kernel solution of Riemannian sparse coding,
followed by
Section~\ref{sec:dic_learning},
which covers the problem of dictionary learning on Riemannian manifolds.
In Section~\ref{sec:experiments}
we compare the performance of the proposed method with previous approaches on several visual classification tasks.
The main findings and possible future directions are summarised in Section~\ref{sec:conclusions}.

\section{Background}
\label{sec:background}

In this section we first overview the properties of Bregman matrix divergences,
including a special case known as the symmetric Stein divergence.
This leads to the Stein kernel,
which can be used to embed Riemannian manifolds into RKHS.

\subsection{Bregman Matrix Divergences}

The {\it Bregman} matrix divergence for two symmetric matrices {\small $\boldsymbol{X}$} and {\small $\boldsymbol{Y}$}
is defined as~\cite{Kulis:2009:JMLR}:

\noindent
\begin{small}
\begin{equation}
    D_{\zeta}(\boldsymbol{X},\boldsymbol{Y}) \triangleq  \zeta( \boldsymbol{X}) - \zeta (\boldsymbol{Y}) -
    \langle \nabla_{\zeta} (\boldsymbol{Y}) , \boldsymbol{X} - \boldsymbol{Y} \rangle
    \label{eqn:Bregman_Div}
\end{equation}%
\end{small}%

\noindent
where
\mbox{\small{$\langle \boldsymbol{A} , \boldsymbol{B} \rangle \mbox{=} \operatorname{Tr} \left( \boldsymbol{A}^T \boldsymbol{B} \right) $}}
and
{\small $\zeta$} is a real valued, strictly convex function on symmetric matrices.
Bregman divergences are non-negative, definite, and in general asymmetric.
Among the several ways to symmetrise them,
the {\it Jensen-Shannon} symmetrisation is often used
\cite{Kulis:2009:JMLR}:

\noindent
\begin{small}
\begin{equation}
    D_{\zeta}^{JS}(\boldsymbol{X},\boldsymbol{Y}) \triangleq  \frac{1}{2}
    D_{\zeta}\left( \boldsymbol{X},\frac{\boldsymbol{X}+\boldsymbol{Y}}{2} \right) +
    \frac{1}{2} D_{\zeta}\left( \boldsymbol{Y},\frac{\boldsymbol{X}+\boldsymbol{Y}}{2} \right)
    \label{eqn:Jensen_Shannon_Div}
\end{equation}%
\end{small}%

\noindent
If {\small $\zeta = - \log \left( \det \left( \boldsymbol{X} \right) \right)$}, then the symmetric Stein divergence is obtained from \eqref{eqn:Jensen_Shannon_Div}:

\noindent
\begin{small}
\begin{equation}
    S(\boldsymbol{X},\boldsymbol{Y}) \triangleq  \log \left( \det \left( \frac{\boldsymbol{X}+\boldsymbol{Y}}{2}\right) \right)
    - \frac{1}{2}  \log \left( \det \left( \boldsymbol{X}\boldsymbol{Y} \right) \right)   ,
    \mbox{~for~} \boldsymbol{X}, \boldsymbol{Y} \succ 0
    \label{eqn:Stein_Div}
\end{equation}%
\end{small}%

The space induced by AIRM on symmetric positive definite matrices of dimension $d$
is a Riemannian manifold {\small $Sym_{+}^{d}$} of negative curvature.
For two points {\small $\boldsymbol{X},\boldsymbol{Y} \in Sym_{+}^{d}$},
the AIRM is defined as
{\small
$
d_g^2
=
\| \log_{\boldsymbol{X}} \left( \boldsymbol{Y} \right)\|_{\boldsymbol{X}}^2
=
\operatorname{Tr} \left\{ \log^2 \left( \boldsymbol{X}^{-\frac{1}{2}} \boldsymbol{Y} \boldsymbol{X}^{-\frac{1}{2}} \right) \right\}
$},
where
{\small
$
\log_{\boldsymbol{X}} \left( \boldsymbol{Y} \right)
=
\boldsymbol{X}^{\frac{1}{2}} \log \left( \boldsymbol{X}^{-\frac{1}{2}} \boldsymbol{Y} \boldsymbol{X}^{-\frac{1}{2}} \right) \boldsymbol{X}^{\frac{1}{2}}
$}.
The symmetric Stein divergence and Riemannian metric over \mbox{$Sym_{+}^{d}$} manifolds are related in several aspects.
Two important properties are summarised below.

\begin{property}
    Let
    {\small $\boldsymbol{X},\boldsymbol{Y} \in Sym_{+}^{d}$},
    and
    {\small $\delta_T(\boldsymbol{X},\boldsymbol{Y}) = \max_{1 \leq i \leq d}
    \{ | \log \Lambda\left(\boldsymbol{X}\boldsymbol{Y}^{-1} \right) |\}$}
    be the Thompson metric~\cite{Thompson:1963}
    with
    {\small $\Lambda\left(\boldsymbol{X}\boldsymbol{Y}^{-1} \right)$}
    representing the vector of eigenvalues of
    {\small $\boldsymbol{X}\boldsymbol{Y}^{-1}$}.
    The following sandwiching inequality between the symmetric Stein divergence and Riemannian metric exists~\cite{Sra:JMLR:2012}:
    
    \noindent
    \begin{small}
    \begin{equation}
        S(\boldsymbol{X},\boldsymbol{Y}) \leq \frac{1}{8} d_g^2(\boldsymbol{X},\boldsymbol{Y}) \leq \frac{1}{4}
        \delta_T(\boldsymbol{X},\boldsymbol{Y})\left( S(\boldsymbol{X},\boldsymbol{Y}) + d\log{d} \right)
        \label{eqn:Sandwich_Inequality}
    \end{equation}%
    \end{small}%

\end{property}

\begin{property}
    The curve
    {\small $\gamma(p) \triangleq \boldsymbol{X}^{\frac{1}{2}} \left(\boldsymbol{X}^{-\frac{1}{2}} \boldsymbol{Y} \boldsymbol{X}^{-\frac{1}{2}} \right)^p \boldsymbol{X}^{\frac{1}{2}}$}
    parameterises the unique geodesic between the SPD matrices {\small $\boldsymbol{X}$} and {\small $\boldsymbol{Y}$}.
    On this curve the Riemannian geodesic distance satisfies
    {\small $d_g(\boldsymbol{X},\gamma(p)) = p d_g(\boldsymbol{X},\boldsymbol{Y}); p \in [0,1]$ }~\cite{BHATIA_2007}.
    The symmetric Stein divergence satisfies a similar but slightly weaker result,
    {\small $S(\boldsymbol{X},\gamma(p)) \leq p S(\boldsymbol{X},\boldsymbol{Y})$}.
\end{property}

The first property establishes a bound between the geodesic distance and Stein divergence,
providing motivation for addressing Riemannian problems via the divergence.
The second property reinforces the motivation
by explaining that the behaviour of Stein divergences
along geodesic curves is similar to true Riemannian geometry.

\subsection{Stein Kernel}
\label{sec:Stein_kernel}

\begin{definition}
Let
{\small $\Omega = \{ \boldsymbol{X}_1,\boldsymbol{X}_2,\cdots,\boldsymbol{X}_N\}$}
be a non-empty set on Riemannian manifold {\small $Sym_{+}^{d}$}.
A function
{\small $\varphi: \Omega \times \Omega \rightarrow \mathbb{R}_+$}
is a Riemannian kernel if {\small $\varphi$} is symmetric for all {\small $\boldsymbol{X}, \boldsymbol{Y} \in \Omega$},
\ie,
{\small $\varphi(\boldsymbol{X}, \boldsymbol{Y}) = \varphi(\boldsymbol{Y},\boldsymbol{X})$},
and the following inequality is satisfied for all \mbox{\small $a_i \in \mathbb{R}$}:
\begin{small}
\begin{equation*}
    \sum \nolimits_{i,j}{a_i a_j \varphi(\boldsymbol{X}_i, \boldsymbol{X}_j)} \geq 0
   \label{eqn:Riemannian_Krnl}
\end{equation*}%
\end{small}%
\end{definition}

\noindent
Under a mild condition (explained afterwards),
the following function forms a Riemannian kernel~\cite{Sra:JMLR:2012}:

\noindent
\begin{small}
\begin{equation}
    k(\boldsymbol{X},\boldsymbol{Y})
    = e^{-\sigma S(\boldsymbol{X},\boldsymbol{Y})}
    = 2^{d \sigma} \frac{\sqrt{\det(\boldsymbol{X})^{\sigma} \det(\boldsymbol{Y})^{\sigma}}}{\det(\boldsymbol{X}+\boldsymbol{Y})^{\sigma}}
    \label{eqn:Stein_Krnl1}
\end{equation}%
\end{small}%

\noindent
We shall refer to this kernel as the {\it Stein kernel} from here on.
The following theorem states the condition under which Stein kernel is positive definite.

\begin{theorem}
Let
{\small $\Omega = \{ \boldsymbol{X}_1,\boldsymbol{X}_2,\cdots,\boldsymbol{X}_N\}; \boldsymbol{X}_i \in Sym_{+}^{d}$}
be a set of Riemannian points.
The {\small $N \times N$} matrix {\small $K_{\sigma} = [k_{\sigma}(i,j)]; 1 \leq i,j \leq N$},
with {\small $k_{\sigma}(i,j)= k(\boldsymbol{X}_i,\boldsymbol{X}_j)$},
defined in~\eqref{eqn:Stein_Krnl1},
is positive definite iff:
\begin{small}
\begin{equation}
    \sigma \in \left \{ \frac{1}{2},\frac{2}{2}, \cdots, \frac{d-1}{2} \right \}
    \cup \left \{\tau \in \mathbb{R}: \tau > \frac{1}{2}(d-1) \right \}
    \label{eqn:Stein_Krnl2}
\end{equation}%
\end{small}%
\end{theorem}

\noindent
Interested readers can follow the proof in~\cite{Sra:JMLR:2012}. For values of $\sigma$ outside of the above set,
it is possible to convert a pseudo kernel into a true kernel,
as discussed for example in~\cite{Similarity_JMLR_2009}.
The determinant of an \mbox{$d \times d$} SPD matrix can be efficiently computed by Cholesky decomposition in \mbox{$O \left( d^3 \right)$}.
As such, the complexity of computing Stein kernel is \mbox{$O\left( 3d^3+3\sigma \right)$}.

\section{Kernel Sparse Coding}
\label{sec:kernel_SR}

Sparse coding on Riemannian manifolds in general means
that a given query point on a manifold can be expressed as a sparse ``combination'' of dictionary elements.
Our idea here is to embed the manifold into RKHS and replace the idea of ``combination'' on manifolds
with the general concept of linear combination in Hilbert spaces.
More specifically, 
given a Riemannian dictionary
{\small $\mathbb{D}=\{\boldsymbol{D}_1,\boldsymbol{D}_2,\cdots,\boldsymbol{D}_N\};{ }\boldsymbol{D}_i \in Sym_{+}^{d}$},
and an embedding function 
{\small $\phi:Sym_{+}^{d} \rightarrow \mathbb{H}$},
for a Riemannian point~{\small $\boldsymbol{X}$}
we seek for a sparse vector {\small $\boldsymbol{v} \in \mathbb{R}^N$}
such that {\small $\phi(\boldsymbol{X})$} admits the sparse representation {\small $\boldsymbol{v}$}
over
{\small $\{\phi(\boldsymbol{D}_1),\phi(\boldsymbol{D}_2),\cdots,\phi(\boldsymbol{D}_N)\}$}.
In other words, we are interested in solving the following kernelised version of the Lasso problem~\cite{ELAD_SR_BOOK_2010}:

\noindent
\begin{small}
\begin{equation}
    \min_{\boldsymbol{v} \in \mathbb{R}^N}
    \left(
    \left\| \phi(\boldsymbol{X}) - \sum\nolimits_{i=1}^{N}v_i \phi(\boldsymbol{D}_i) \right\|^2
    +
    \lambda \left\|\boldsymbol{v} \right\|_1
    \right)
    \label{eqn:KSR_Opt1}
\end{equation}%
\end{small}%

\noindent
The first term in~\eqref{eqn:KSR_Opt1} can be expanded as:
\begin{small}
\begin{align}
    & \left\| \phi(\boldsymbol{X})-\sum\nolimits_{i=1}^{N}v_i \phi(\boldsymbol{D}_i) \right\|^2 \nonumber \\
    & = k(\boldsymbol{X},\boldsymbol{X})-2\sum\nolimits_{i=1}^{N}{v_i k(\boldsymbol{X},\boldsymbol{D}_i)}+\sum\nolimits_{i=1}^{N}{\sum\nolimits_{j=1}^{N}{v_i v_j k(\boldsymbol{D}_j,\boldsymbol{D}_i)}} \nonumber \\
    & = k(\boldsymbol{X},\boldsymbol{X})-2\boldsymbol{v}^T\boldsymbol{\mathcal{K}}(\boldsymbol{X},\mathbb{D})+\boldsymbol{v}^T \boldsymbol{\mathbb{K}(\mathbb{D},\mathbb{D})} \boldsymbol{v}
    \label{eqn:KSR_Opt2}
\end{align}%
\end{small}%

\noindent
where
{\small $\boldsymbol{\mathcal{K}}=[a_{i}]_{N \times 1}; ~ a_{i}=k(\boldsymbol{X},\boldsymbol{D}_i)$}
and
{\small $\boldsymbol{\mathbb{K}}=[a_{ij}]_{N \times N}; ~ a_{ij}=k(\boldsymbol{D}_i,\boldsymbol{D}_j)$}.
This reveals that the optimisation problem in \eqref{eqn:KSR_Opt1}
is convex and similar to its counterpart in Euclidean space,
except for the definition of {\small $\boldsymbol{\mathcal{K}}$} and {\small $\boldsymbol{\mathbb{K}}$}.
Consequently, greedy or relaxation solutions can be adapted to obtain the sparse codes~\cite{ELAD_SR_BOOK_2010}.
To solve problem \eqref{eqn:KSR_Opt1} we used CVX~\cite{CVX},
a package for specifying and solving convex programs\footnote{The SPAMS package can also be used: \url{http://spams-devel.gforge.inria.fr/}}.

\subsection{Classification Using Sparse Codes}

There are two main approaches for classification
based on the obtained sparse codes (vectors) for a given query sample:
{\bf (i)}~directly,
and
{\bf (ii)}~indirectly, with the aid of an Euclidean-based classifier.
We elucidate the two approaches below.

{\bf (i)}~If~the atoms in sparse dictionary {\small $\mathbb{D}$} have associated class labels
(ie.~each atom in the dictionary is a training sample),
the sparse codes can be directly used for classification.
This approach is applicable only to closed-set identification tasks.
Let
{\small $\boldsymbol{v}_i = [ v_{i,1}\delta( l(1)-i ), ~v_{i,2}\delta( l(2)-i ), ~\cdots, ~v_{i,N}\delta( l(N)-i ) ]^T$}
be the class-specific sparse codes,
where {\small $l(j)$} is the class label of atom~{\small $\boldsymbol{D}_j$}
and {\small $\delta(x)$} is the discrete Dirac function~\cite{Bishop_2006}.
An efficient way of using class-specific sparse codes is through computing residual errors~\cite{Wright_2009_PAMI}.
In this case, the residual error of query sample {\small $\boldsymbol{X}$} for class {\small $i$} is defined as:

\noindent
\begin{small}
\begin{equation}
    \varepsilon_i(\boldsymbol{X})= \left\| \boldsymbol{\phi(X)} - \sum\nolimits_{j=1}^{N} v_{j} \phi(\boldsymbol{D}_j) \delta( l(j)-i ) \right\|^2
    \label{eqn:sparse_classification1}
\end{equation}%
\end{small}%

\noindent
which can be computed via the use of a Riemannian kernel in a similar manner to~\eqref{eqn:KSR_Opt2}.
The class with the minimum residual error is deemed to represent the query.
Alternatively,
the similarity between query sample {\small $\boldsymbol{X}$} to class {\small $i$}
can be defined as \mbox{\small $S_i(\boldsymbol{X}) \mbox{=} h_{i}(\boldsymbol{v})$}.
The function
{\small $h_{i}(\boldsymbol{v})$} can be linear like \mbox{\small $\sum\nolimits_{j=1}^{j=N} v_{j}\delta( l(j)-i )$}
or even non-linear like
\linebreak
{\small $\max\left(v_{j}\delta( l(j)-i )\right)$}.

{\bf (ii)}~If~the atoms in the sparse dictionary {\small $\mathbb{D}$} are not labelled
(eg.~{\small $\mathbb{D}$} is a generic dictionary not tied to any particular class~\cite{Yong_IJCNN_2012}),
the generated sparse codes (vectors) for both training and query data can be fed to Euclidean-based classifiers,
such as support vector machines~\cite{Bishop_2006}.
The sparse code is hence interpreted as a feature vector,
which in essence means that
a classification problem on a Riemannian manifold
is converted to an Euclidean classification problem.
This approach is applicable to both closed-set and open-set classification tasks.

\section{Learning Riemannian Dictionaries}
\label{sec:dic_learning}

If the indirect classification of sparse codes is required (as elucidated in the preceding section)
a Riemannian dictionary is first required.
Given a set of Riemannian points
{\small $\Omega = \{ \boldsymbol{X}_1,\boldsymbol{X}_2,\cdots,\boldsymbol{X}_m\}; \boldsymbol{X}_i \in Sym_{+}^{d}$},
learning a dictionary
{\small $\mathbb{D} = \{ \boldsymbol{D}_1,\boldsymbol{D}_2,\cdots,\boldsymbol{D}_N\}; \boldsymbol{D}_i \in Sym_{+}^{d}$}
can be formulated as jointly minimising the energy function

\noindent
\begin{small}
\begin{equation}
     J = \left\{
    \sum \nolimits_{j = 1}^{m}{ \left( \left\| \phi(\boldsymbol{X}_j) - \sum \nolimits_{i=1}^{N}v_{j,i} \phi(\boldsymbol{D}_i) \right\|^2
    +     \lambda \left\|\boldsymbol{v}_j \right\|_1 \right)}\right\}
    \label{eqn:dic_lrn1}
\end{equation}%
\end{small}%

\noindent
over the dictionary and the sparse codes
{\small $\mathbb{V} = \{\boldsymbol{v}_1,\boldsymbol{v}_2, \cdots, \boldsymbol{v}_m\}; \boldsymbol{v}_i \in \mathbb{R}^N$},
\ie,
{\small ${\min}_{_{\mathbb{D},\mathbb{V}}}(J)$}.

Among the various solutions to the problem of dictionary learning in Euclidean spaces, iterative methods
like K-SVD have received much attention~\cite{ELAD_SR_BOOK_2010}.
Borrowing the idea from Euclidean spaces,
we propose to minimise the energy in \eqref{eqn:dic_lrn1} iteratively.
After initialising the dictionary $\mathbb{D}$,
for example by Riemannian clustering using the Karcher mean~\cite{Pennec_jmiv06},
we iterate between a sparse coding step and a dictionary update step.
In the sparse coding step,
$\mathbb{D}$ is fixed and $\mathbb{V}$ is computed.
In the dictionary update step,
$\mathbb{V}$ is fixed while $\mathbb{D}$ is updated,
with each dictionary atom updated independently.

The derivative of \eqref{eqn:dic_lrn1} with respect to \mbox{\small{$\boldsymbol{D}_r$}},
while $\mathbb{V}$ and other atoms are fixed, is:
\begin{small}
\begin{equation}
    \frac{\partial J}{\partial \boldsymbol{D}_r} =
    \sum \nolimits_{j=1}^{m} \left( -2 v_{j,r} \frac{\partial k(\boldsymbol{X}_j,\boldsymbol{D}_r )}{\partial \boldsymbol{D}_r}
    + \sum \nolimits_{i=1}^{N} v_{j,i} v_{j,r} \frac{\partial k(\boldsymbol{D}_i,\boldsymbol{D}_r)}{\partial \boldsymbol{D}_r}  \right)
    \label{eqn:dic_lrn2}
\end{equation}%
\end{small}%

\noindent
As
{\small $\nabla_{\boldsymbol{X}} S(\boldsymbol{X},\boldsymbol{Y}) = (\boldsymbol{X}+\boldsymbol{Y})^{-1} - \frac{1}{2} \boldsymbol{X}^{-1}$},
\eqref{eqn:dic_lrn2} can be further simplified to:
\begin{small}
\begin{align}
    \frac{\partial J}{\partial \boldsymbol{D}_r}
    =
    &
    ~ ~ ~ 2 \beta \sum \nolimits_{j=1}^{m}
    { v_{j,r} k(\boldsymbol{X}_j,\boldsymbol{D}_r ) \left( (\boldsymbol{X}_j+\boldsymbol{D}_r)^{-1} -\frac{1}{2} \boldsymbol{D}_r^{-1} \right)} \nonumber \\
    &
    - \beta \sum \nolimits_{j=1}^{m} {\sum \nolimits_{i=1}^{N} {v_{j,i} v_{j,r} k(\boldsymbol{D}_i,\boldsymbol{D}_r ) \left( (\boldsymbol{D}_i+\boldsymbol{D}_r)^{-1}
    -\frac{1}{2} \boldsymbol{D}_r^{-1} \right)}}
    \label{eqn:dic_lrn3}
\end{align}%
\end{small}%

\noindent
Since \eqref{eqn:dic_lrn3} contains linear and non-linear terms of {\small $\boldsymbol{D}_r$}
(\eg inverse and kernel terms),
a~closed-form solution for computing its root,
\ie, {\small $\boldsymbol{D}_r$}, cannot be sought.
As such, we propose an alternative solution
by exploiting previous values of
{\small $k(\cdot,\boldsymbol{D}_r)$}
and
{\small $ \left ( \boldsymbol{D}_i - \boldsymbol{D}_r \right )^{-1}$}
in the updating step.
More specifically, rearranging \eqref{eqn:dic_lrn3} and estimating
{\small $k(\cdot,\boldsymbol{D}_r)$}
as well as
{\small $ \left ( \boldsymbol{D}_i - \boldsymbol{D}_r \right )^{-1}$}
by their previous values,
atom {\small $\boldsymbol{D}_r$} at iteration {\small $t+1$} is updated using:

\noindent
\begin{small}
\begin{equation}
    \boldsymbol{D}_r^{(t+1)} =
    \frac{2}{\sum \nolimits_{j=1}^{m}{v_{j,r}
     \left( \boldsymbol{v}_{j}^{T} \boldsymbol{k}(\mathbb{D},\boldsymbol{D}_r) -2k(\boldsymbol{X}_j,\boldsymbol{D}_r) \right)}}
    \left ( F^{(t)}(r) + G^{(t)}(r) \right ) ^{-1}
    \label{eqn:dic_lrn4}
\end{equation}%
\end{small}%

\noindent
where
\begin{small}
\begin{eqnarray}
    F^{(t)}(r) & = & \sum \nolimits_{j=1}^{m}{2v_{j,r} k^{(t)}(\boldsymbol{X}_j,\boldsymbol{D}_r)\left( \boldsymbol{X}_j + \boldsymbol{D}_r^{(t)} \right)^{-1}}
    \label{eqn:dic_lrn5}
    \\
    G^{(t)}(r) & = & \sum \nolimits_{j=1}^{m}{\sum \nolimits_{i=1}^{N}{v_{j,r}v_{j,i} k^{(t)}(\boldsymbol{D}_i,\boldsymbol{D}_r)
    \left( \boldsymbol{D}_i^{(t)} + \boldsymbol{D}_r^{(t)} \right)^{-1}}}
    \label{eqn:dic_lrn6}
\end{eqnarray}%
\end{small}%

To avoid the degenerative case (due to numerical inconsistency), atoms are normalised by their second norms at the end of each iteration.
Algorithm~\ref{alg:dic_learning_pseudo_code} assembles all the above details into pseudo-code for dictionary learning.

\begin{algorithm}[!tb]
\caption{: Dictionary learning over {\small $Sym_{+}^{d}$} using the Stein kernel}
\label{alg:dic_learning_pseudo_code}
\begin{algorithmic}[1]

\REQUIRE
~\\
\vspace{-2ex}
\begin{itemize}

\item
training set {$\mathbb{X} \mbox{=} \left\{  \boldsymbol{X}_i \right\}_{i=1}^{m}$} from the underlying Riemannian manifold,\\
where each {$\boldsymbol{X}_i \in Sym_{+}^{d}$} is a SPD matrix

\item
Stein kernel function
{\small $k(\boldsymbol{X},\boldsymbol{Y})$}, as defined in Eqn.~\eqref{eqn:Stein_Krnl1}

\item
$nIter$, the number of iterations

\end{itemize}

\ENSURE
~\\
\vspace{-2ex}
\begin{itemize}

\item
Riemannian dictionary \mbox{$\mathbb{D}= \left\{  \boldsymbol{D}_i \right\}_{i=1}^{N}$}

\end{itemize}

\noindent

\STATE Initialise the dictionary
{$\mathbb{D}^{(1)}= \left\{  \boldsymbol{D}_i^{_{(1)}} \right\}_{i=1}^{N}$}
by selecting $N$ samples from $\mathbb{X}$ randomly,\\
or
by clustering $\mathbb{X}$ on the manifold~\cite{VIDAL:CVPR:2008}

~

\FOR{ $t = 1 \to nIter$}

    \STATE Compute \mbox{$k^{_{(t)}}(\boldsymbol{X}_i,\boldsymbol{D}_j^{_{(t)}}), ~ 1\leq i \leq m, ~ 1\leq j \leq N$}

    \STATE Compute \mbox{$k^{_{(t)}}(\boldsymbol{D}_i^{_{(t)}},\boldsymbol{D}_j^{_{(t)}}), ~ 1\leq i,j \leq N$}

\STATE
Solve
{$\underset{\boldsymbol{v}_j \in \mathbb{R}^N}{\min}
         - 2 \boldsymbol{v}_j^T\boldsymbol{\mathcal{K}}(\boldsymbol{X}_j,\mathbb{D}^{(t)})
         + \boldsymbol{v}^T \mathbb{K}(\mathbb{D}^{(t)},\mathbb{D}^{(t)}) \boldsymbol{v}
         + \lambda \left\|\boldsymbol{v}_j \right\|_1 $}, ~
{$\forall j, \boldsymbol{X}_j \in \mathbb{X}$}

\FOR{ $r = 1 \to N$}

\STATE
Compute
\mbox{$G^{(t)}(r) = \sum_{j=1}^{m}{\sum_{i=1}^{N}{v_{j,r} v_{j,i} k^{(t)}(\boldsymbol{D}_i^{(t)},\boldsymbol{D}_r^{(t)})
    \left( \boldsymbol{D}_i^{(t)} + \boldsymbol{D}_r^{(t)} \right)^{-1}}}$}

\STATE
Compute
\mbox{$F^{(t)}(r) = \sum_{j=1}^{m}{2v_{j,r} k^{(t)}(\boldsymbol{X}_j,\boldsymbol{D}_r^{(t)})\left( \boldsymbol{X}_j + \boldsymbol{D}_r^{(t)} \right)^{-1}}$}

\STATE
\mbox{$
\boldsymbol{D}_r^{_{\mathtt{temp}}} \gets
    \frac{2}{\sum_{j=1}^{m}{v_{j,r}
     \left( \boldsymbol{v}_{j}^{T} \boldsymbol{k}(\mathbb{D}^{(t)},\boldsymbol{D}_r^{(t)}) -2k(\boldsymbol{X}_j,\boldsymbol{D}_r^{(t)}) \right)}}
\left ( F^{(t)}(r) + G^{(t)}(r) \right ) ^{-1}
$}

\STATE
\mbox{$
\boldsymbol{D}_r^{_{(t+1)}} \gets \frac{\boldsymbol{D}_r^{_{ \mathtt{temp} }}}{\left\| \boldsymbol{D}_r^{_{ \mathtt{temp} }} \right\|_2}
$}
\ENDFOR
\ENDFOR
\end{algorithmic}
\end{algorithm}

\section{Experiments}
\label{sec:experiments}

Two sets of experiments\footnote{Matlab/Octave source code is available at \mbox{\url{http://itee.uq.edu.au/~uqmhara1}}}
are presented in this section.
In the first set,
we evaluate the performance of the proposed Riemannian SR (RSR) method (as described in Section~\ref{sec:kernel_SR})
without dictionary learning.
Each atom in the dictionary is a training sample.
This is to contrast RSR to previous state-of-the-art methods on several popular closed-set classification tasks.
We use the residual error approach for classification,
as described in Eqn.~\eqref{eqn:sparse_classification1}.

In the second set, the performance of the RSR method is evaluated
in conjunction with dictionaries learned via three methods:
random, Riemannian {\it k}-means,
and the proposed dictionary learning technique
(as described in Section~\ref{sec:dic_learning}).

\subsection{Riemannian Sparse Representation}
\label{sec:exp_rsr_closed_set}

\subsubsection{Synthetic Data.}
\label{sec:exp_syn_data}

We first consider a multi-class classification problem over {\small $Sym_+^3$} using synthetic data.
We compared the proposed RSR against Tensor Sparse Coding  (TSC)~\cite{Sivalingam:ECCV:2010}
and log-Euclidean Sparse Representation (logE-SR)~\cite{SR_Riemannian_AVSS_2010,Yuan:ACCV:2010}.
The data used in the experiments constitutes 512 random samples from 4 classes.
Half of the samples were used for training and the rest were used as test data.

To create a Riemannian manifold, samples were generated over a particular tangent space
and then mapped back to the manifold using the exponential map~\cite{BHATIA_2007}.
The positions of tangent spaces were chosen randomly
and samples in each class obeyed a normal distribution.
By fixing the mean of each class and increasing the class variance
we created two classification problems: `easy' and `hard'.
To draw useful statistics, the data creation process was repeated 100 times.

Table~\ref{tab:table_toy_data} shows the average recognition accuracy and the total running time (in seconds).
All algorithms were implemented in Matlab and executed on a 3 GHz Intel CPU.
In terms of recognition accuracy, RSR obtains superior performance when compared with previous state-of-the-art approaches.
We note that by increasing the class variance,
samples from the four classes are intertwined,
leading to a decrease in recognition accuracy.
The performance of logE-SR is higher than TSC,
which might be due the to fact that the generated data can be modelled
by Gaussian distribution over tangent space, hence favouring the tangent-based solution.

Focusing on run time,
Table~\ref{tab:table_toy_data} suggests that logE-SR has the lowest complexity while TSC has the highest.
The proposed RSR method is substantially faster than TSC,
while delivering the highest recognition accuracy.

\begin{table}[!t]
    \centering
    \begin{tabular}{rlcccc}
      \toprule
      ~
      &
      &
      &{\bf logE-SR~\cite{SR_Riemannian_AVSS_2010,Yuan:ACCV:2010}}
      &~~~~~~{\bf TSC~\cite{Sivalingam:ECCV:2010}}~~~~~~
      &{\bf RSR (proposed)}\\
      \toprule
       \multirow{4}{*}{\rotatebox{90}{\fontsize{8}{1.0}\selectfont recognition}}
      &\multirow{4}{*}{\rotatebox{90}{\fontsize{8}{1.0}\selectfont accuracy}}
      \\
      & &{\bf easy}    &$68.08 \pm 2.5$ &$60.04 \pm 6.8$ &${\bf 83.05 \pm 3.0}$ \\
      & &{\bf hard}    &$53.67 \pm 3.2$ &$50.35 \pm 4.9$ &${\bf 66.72 \pm 2.7}$ \\
      \\
     \hline
     \multirow{3}{*}{\rotatebox{90}{\fontsize{8}{1.0}\selectfont run-time}}
     &\multirow{3}{*}{\rotatebox{90}{\fontsize{8}{1.0}\selectfont ~}}
     \\
     & &{\bf ~easy+hard}    &$6$ sec   &$11107$ sec & $41$ sec \\
     \\
      \bottomrule
    \end{tabular}

    ~

    \caption
      {
      Average recognition accuracy (in \%) and wall-clock time for the synthetic classification tasks
      using log-Euclidean sparse representation~\cite{SR_Riemannian_AVSS_2010,Yuan:ACCV:2010},
      tensor sparse coding~\cite{Sivalingam:ECCV:2010} and the proposed RSR approach.
      Run time is represented by combining the times for the easy and hard tasks.
      }
      \label{tab:table_toy_data}
\end{table}

\subsubsection{Face Recognition.}
\label{sec:exp_face}

\begin{figure}[!b]
  \centering
  \begin{minipage}{1.0\columnwidth}
  \centering
      \begin{minipage}{0.15\columnwidth}
          \centerline{\includegraphics[scale = 0.75]{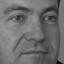}}
          \centerline{\footnotesize bd}
        \end{minipage}
      \begin{minipage}{0.15\columnwidth}
          \centerline{\includegraphics[scale = 0.75]{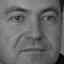}}
          \centerline{\footnotesize be}
        \end{minipage}
      \begin{minipage}{0.15\columnwidth}
          \centerline{\includegraphics[scale = 0.75]{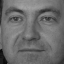}}
          \centerline{\footnotesize bf}
        \end{minipage}
      \begin{minipage}{0.15\columnwidth}
          \centerline{\includegraphics[scale = 0.75]{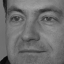}}
          \centerline{\footnotesize bg}
        \end{minipage}
      \begin{minipage}{0.15\columnwidth}
          \centerline{\includegraphics[scale = 0.75]{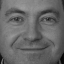}}
          \centerline{\footnotesize bj}
        \end{minipage}
      \begin{minipage}{0.15\columnwidth}
          \centerline{\includegraphics[scale = 0.75]{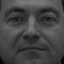}}
          \centerline{\footnotesize bk}
        \end{minipage}
  \end{minipage}
  \caption
    {
    \small
    Examples of closely-cropped faces from the FERET `b' subset.
    }
  \label{fig:feret_sample}
\end{figure}

We used the `b' subset of the FERET dataset~\cite{FERET_DATASET},
which includes 1400 images from 198 subjects.
The images were closely cropped around the face and downsampled to \mbox{\small $64 \times 64$}.
Examples are shown in Figure~\ref{fig:feret_sample}.

We performed four tests with various pose angles.
Training data was composed of images marked `ba', `bj' and `bk' (\ie, frontal faces with expression and illumination variations).
Images with `bd', `be', `bf' and  `bg' labels (\ie, \mbox{non-frontal} faces) were used as test data.
For Riemannian-based methods,
a \mbox{$43 \times 43$} covariance descriptor described a face image,
using the following features:

\noindent
\begin{small}
\begin{equation*}
F_{x,y}
\mbox{=}
\left[~ I(x,y),~ x,~ y,~ |G_{0,0}(x,y)|,~ \cdots\hspace{-0.4ex},~ |G_{0,7}(x,y)|,~ |G_{1,0}(x,y)|,~ \cdots\hspace{-0.4ex},~ |G_{4,7}(x,y)| ~\right]
\end{equation*}%
\end{small}%

\noindent
where {\small $I(x,y)$} is the intensity value at position {\small $x,y$}
and
{\small $G_{u,v}{(x,y)}$}
is the response of a 2D Gabor wavelet~\cite{Lee:1996:PAMI}
centered at {\small $x,y$} with orientation {\small $u$} and scale {\small $v$}:
\begin{small}
\begin{equation*}
    G_{u,v}(x,y) = \frac{k_v^2}{4\pi^2} \sum\nolimits_{t,s}
                    e^{ - \frac{k_v^2}{8\pi^2} \left((x-s)^2+(y-t)^2 \right)}
                    \left ( e^{i k_v \left( (x-t)cos(\theta_u) + (y-s)sin(\theta_u) \right)}
                    - e^{-2\pi^2}\right )
\end{equation*}%
\end{small}%

\noindent
with \mbox{\small{$k_v =\frac{1}{\sqrt{2^{v-1}}}$}} and \mbox{\small{$\theta_u = \frac{\pi u}{8}$}}.

Table~\ref{tab:table_face_recognition} shows a comparison of RSR
against
logE-SR~\cite{SR_Riemannian_AVSS_2010,Yuan:ACCV:2010},
TSC~\cite{Sivalingam:ECCV:2010},
and two purely Euclidean sparse representations,
PCA-SRC~\cite{Wright_2009_PAMI}
and Gabor SR (GSR)~\cite{GSR:ECCV2010}.
In all cases the proposed RSR method obtains the highest accuracy.
Furthermore, the proposed approach significantly outperforms state-of-the-art Euclidean solutions,
especially for test images with label `bg'.

\begin{table}[!t]
  \centering
    \begin{tabular}{lccccc}
    \toprule
    {\bf }  &~~~{\bf PCA-SR~\cite{Wright_2009_PAMI}}~~~
    &~~~{\bf GSR~\cite{GSR:ECCV2010}}~~~
    &~~~{\bf logE-SR~\cite{SR_Riemannian_AVSS_2010,Yuan:ACCV:2010}}~~~
    &~~~{\bf TSC~\cite{Sivalingam:ECCV:2010}}~~~
    &{\bf RSR (proposed)} \\
    \toprule
    {\bf bg}     &$26.0$      &$79.0$      &$46.5$   &$44.5$      &${\bf 86.0}$   \\
    {\bf bf}     &$61.0$      &$97.0$      &$91.0$   &$73.5$      &${\bf 97.5}$   \\
    {\bf be}     &$55.5$      &$93.5$      &$81.0$   &$73.0$      &${\bf 96.5}$   \\
    {\bf bd}     &$27.5$      &$77.0$      &$34.5$   &$36.0$      &${\bf 79.5}$   \\
    {\bf average}&$42.50$     &$86.63$     &$63.25$  &$56.75$     &${\bf 89.88}$  \\
    \bottomrule
    \end{tabular}

  ~

  \caption
    {
    Recognition accuracy (in \%) for the face recognition task
    using
    PCA-SRC~\cite{Wright_2009_PAMI},
    Gabor SR (GSR)~\cite{GSR:ECCV2010},
    log-Euclidean sparse representation (logE-SR)~\cite{SR_Riemannian_AVSS_2010,Yuan:ACCV:2010},
    Tensor Sparse Coding (TSC)~\cite{Sivalingam:ECCV:2010},
    and the proposed RSR approach.
    }
    \label{tab:table_face_recognition}
\end{table}

\subsubsection{Texture Classification.}
\label{sec:exp_texture}

\begin{figure}[!t]
  \begin{minipage}{1\textwidth}
    \begin{minipage}{0.25\textwidth}
       \centering
       \includegraphics[width=0.5\textwidth,keepaspectratio]{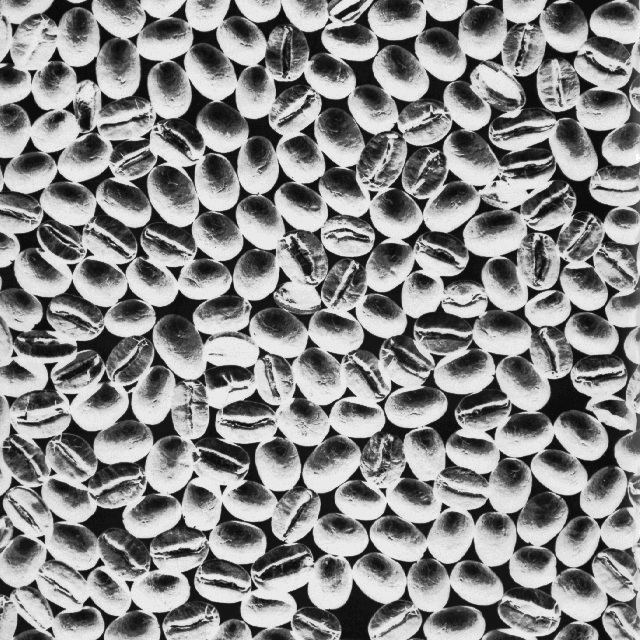}\\

       ~

       \includegraphics[width=0.5\textwidth,keepaspectratio]{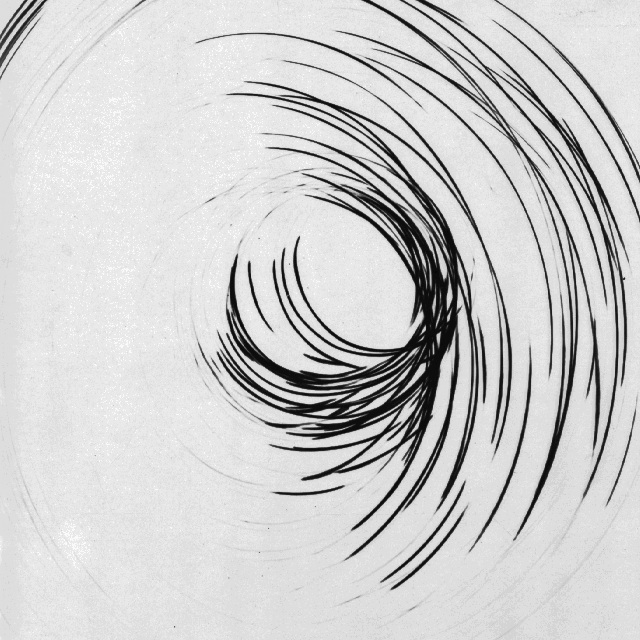}\\

       ~

       \includegraphics[width=0.5\textwidth,keepaspectratio]{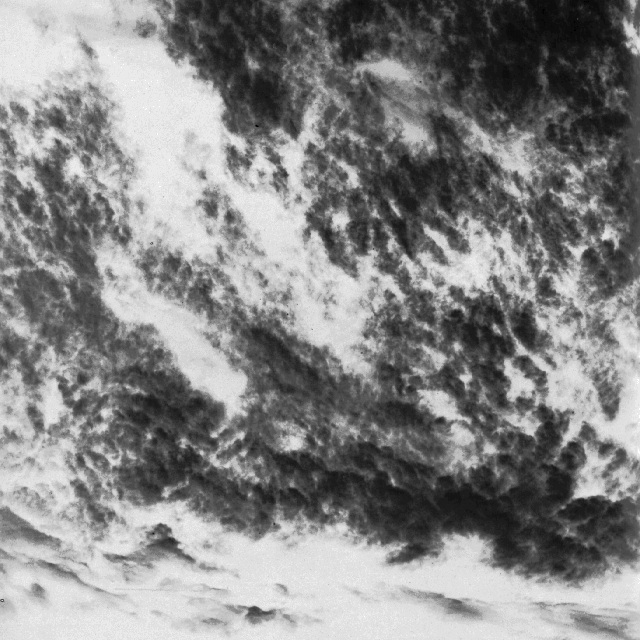}
       \caption
        {
        \small
        Examples from the Brodatz texture dataset~\cite{Brodatz_Dataset}.
        }
      \label{fig:Texture_example}
    \end{minipage}
    \hfill
    \begin{minipage}{0.7\textwidth}
      \begin{minipage}{0.05\textwidth}
        \rotatebox[origin=l]{90}{\scriptsize Average recognition accuracy}
      \end{minipage}
      \begin{minipage}{0.95\textwidth}
        \centering
        \includegraphics[width=\textwidth,keepaspectratio]{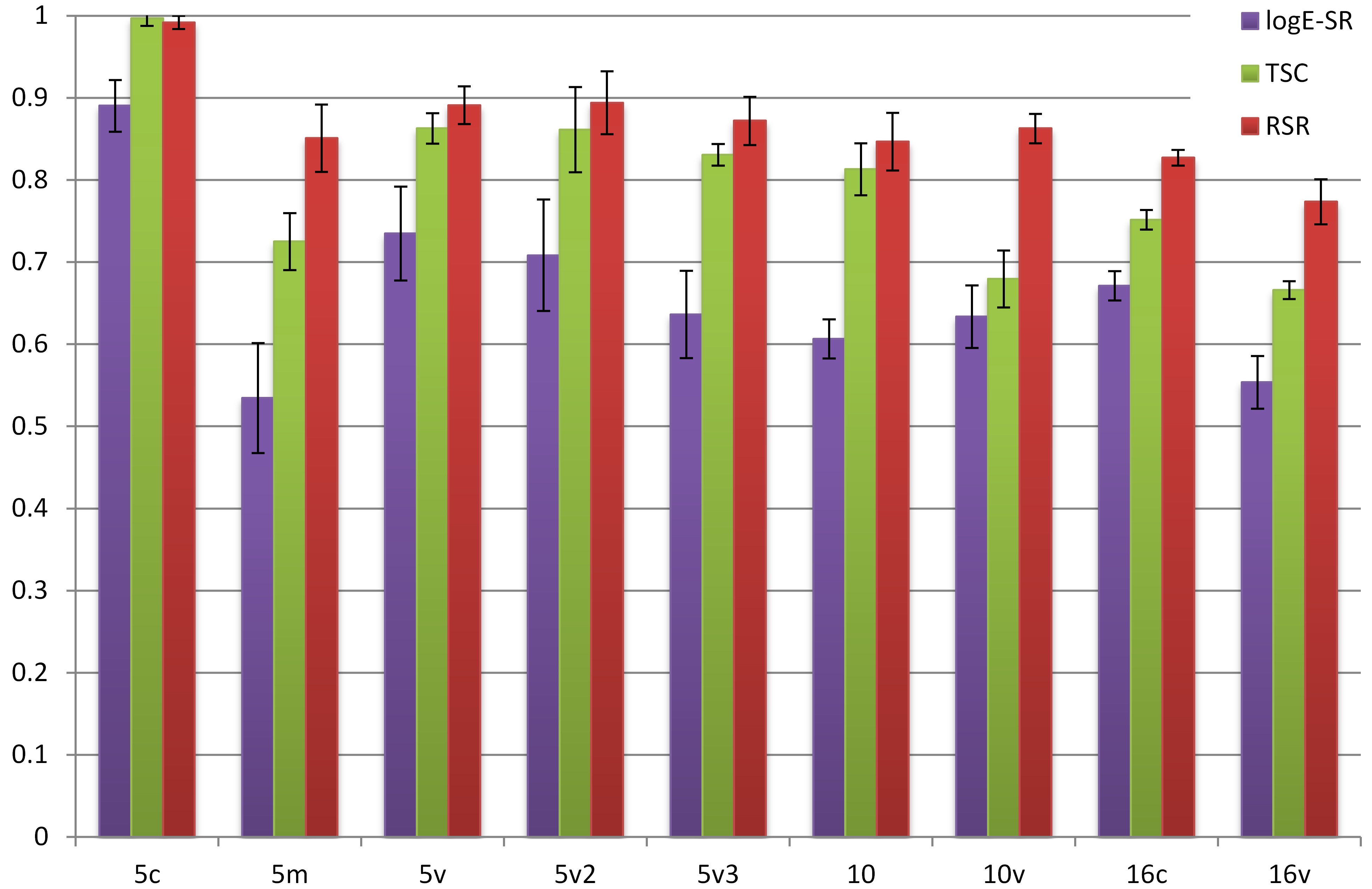}
        {\scriptsize Test ID}
      \end{minipage}
      \caption
        {
        \small
        Performance on the Bordatz texture dataset~\cite{Brodatz_Dataset} using
        log-Euclidean sparse representation (logE-SR)~\cite{SR_Riemannian_AVSS_2010,Yuan:ACCV:2010},
        Tensor Sparse Coding (TSC)~\cite{Sivalingam:ECCV:2010}
        and
        the proposed RSR approach.
        The black bars indicate standard deviations.
        }
    \label{fig:Texture_results}
    \end{minipage}
  \end{minipage}
\end{figure}

We performed a classification task using the Brodatz texture dataset~\cite{Brodatz_Dataset}.
Examples are shown in Fig.~\ref{fig:Texture_example}.
We followed the test protocol devised in~\cite{Sivalingam:ECCV:2010}
and generated nine test scenarios with various number of classes.
This includes 5-texture (`5c', `5m', `5v', `5v2', `5v3'),
10-texture (`10', `10v') and 16-texture (`16c', `16v') mosaics.
To create a Riemannian manifold,
each image was first downsampled to {\small $256 \times 256$}
and then split into 64 regions of size {\small $32 \times 32$}.
The feature vector for any pixel {\small $I\left(x,y\right)$}
is
\mbox
  {\footnotesize
  $F(x, y)
  \mbox{=}
  \left[
    I\left(x,y\right),
    \left| \frac{\partial I}  {\partial x}  \right|,  \left|\frac{\partial I}  {\partial y}  \right|,
    \left| \frac{\partial^2 I}{\partial x^2}\right|,  \left|\frac{\partial^2 I}{\partial y^2}\right|
  \right]$}.
Each region is described by a {\small $5 \times 5$} covariance descriptor of these features.
For each test scenario, five covariance matrices per class were randomly selected as training data and the rest was used for testing.
The random selection of training/testing data was repeated 20 times.

Fig.~\ref{fig:Texture_results} compares the proposed RSR method against
logE-SR~\cite{SR_Riemannian_AVSS_2010,Yuan:ACCV:2010}
and TSC~\cite{Sivalingam:ECCV:2010}.
The proposed RSR approach obtains the highest recognition accuracy on all test scenarios
except for the `5c' test,
where it has slightly worse performance than TSC.

\begin{figure}[!t]
  \begin{minipage}{1\textwidth}
    \begin{minipage}{0.6\textwidth}
      \centering
      \includegraphics[width=\textwidth,keepaspectratio]{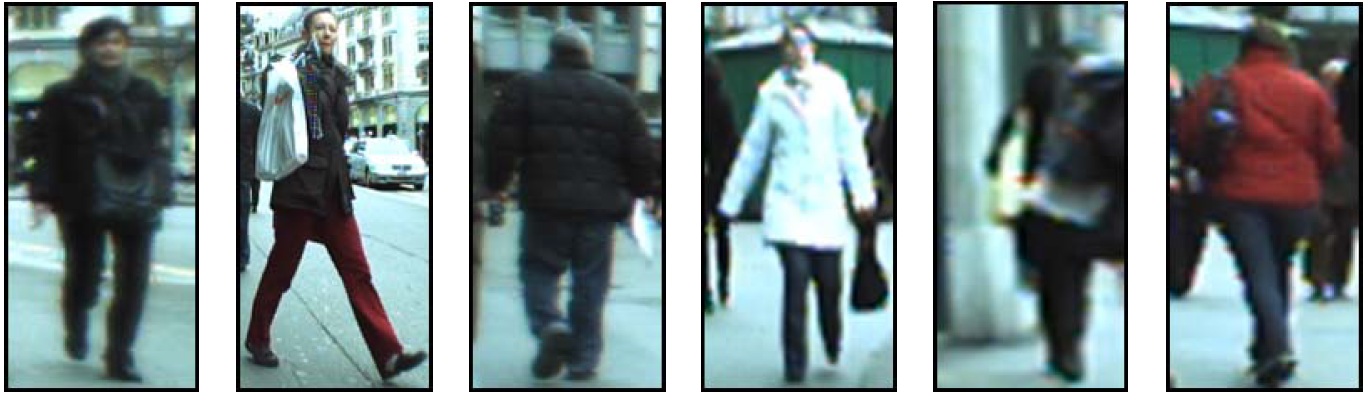}
    \end{minipage}
    \hfill
    \begin{minipage}{0.35\textwidth}
      \caption
        {
        Examples of pedestrians in the ETHZ dataset~\cite{ETHZ_ICCV}.
        }
      \label{fig:ETHZ_examples}
    \end{minipage}
  \end{minipage}
\end{figure}

\begin{figure}[!t]
    \begin{minipage}{1.0\textwidth}
      \begin{minipage}{0.5\textwidth}
        \centering
        \includegraphics[width=\textwidth,keepaspectratio]{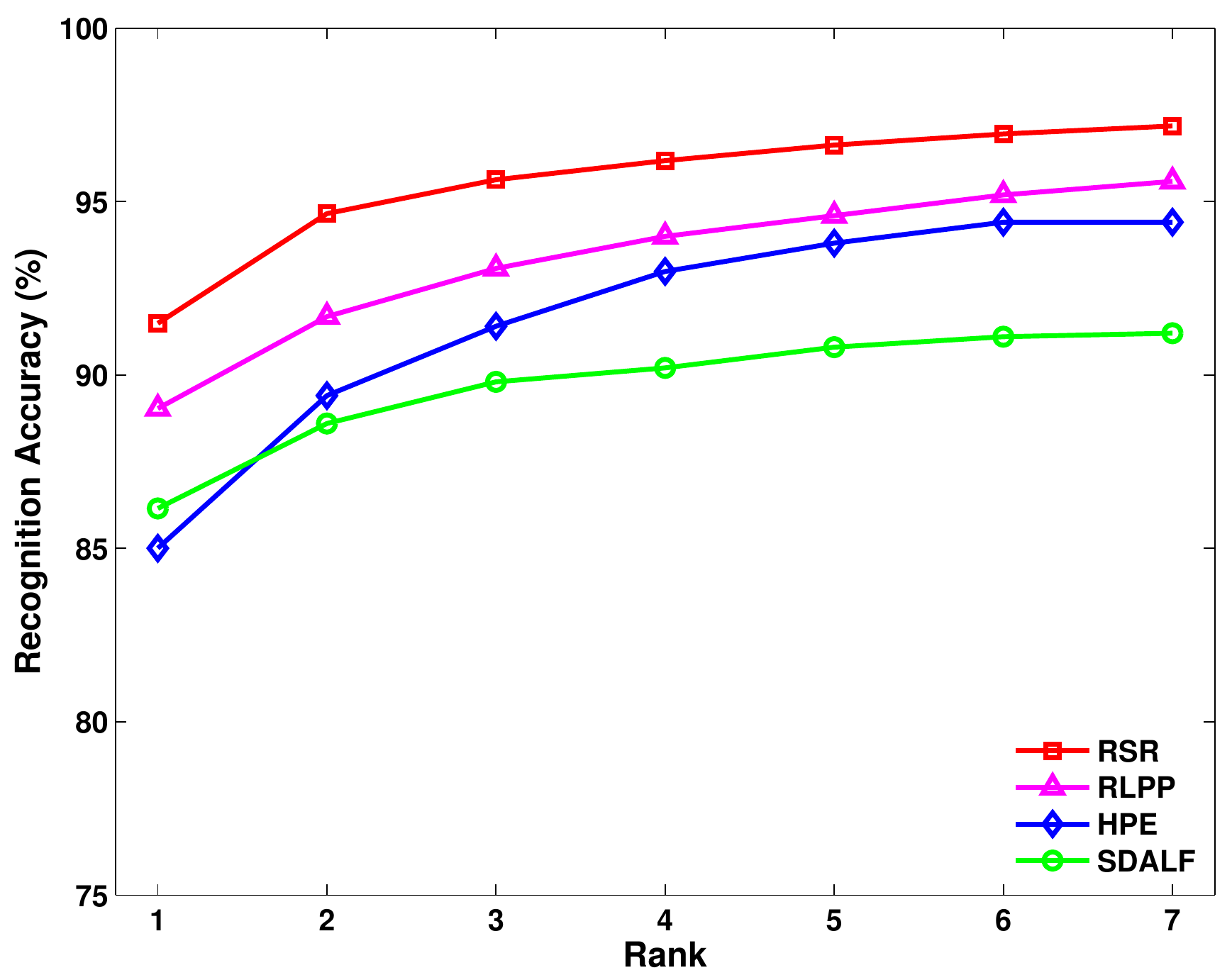}\\
      \end{minipage}
      \begin{minipage}{0.5\textwidth}
        \centering
        \includegraphics[width=\textwidth,keepaspectratio]{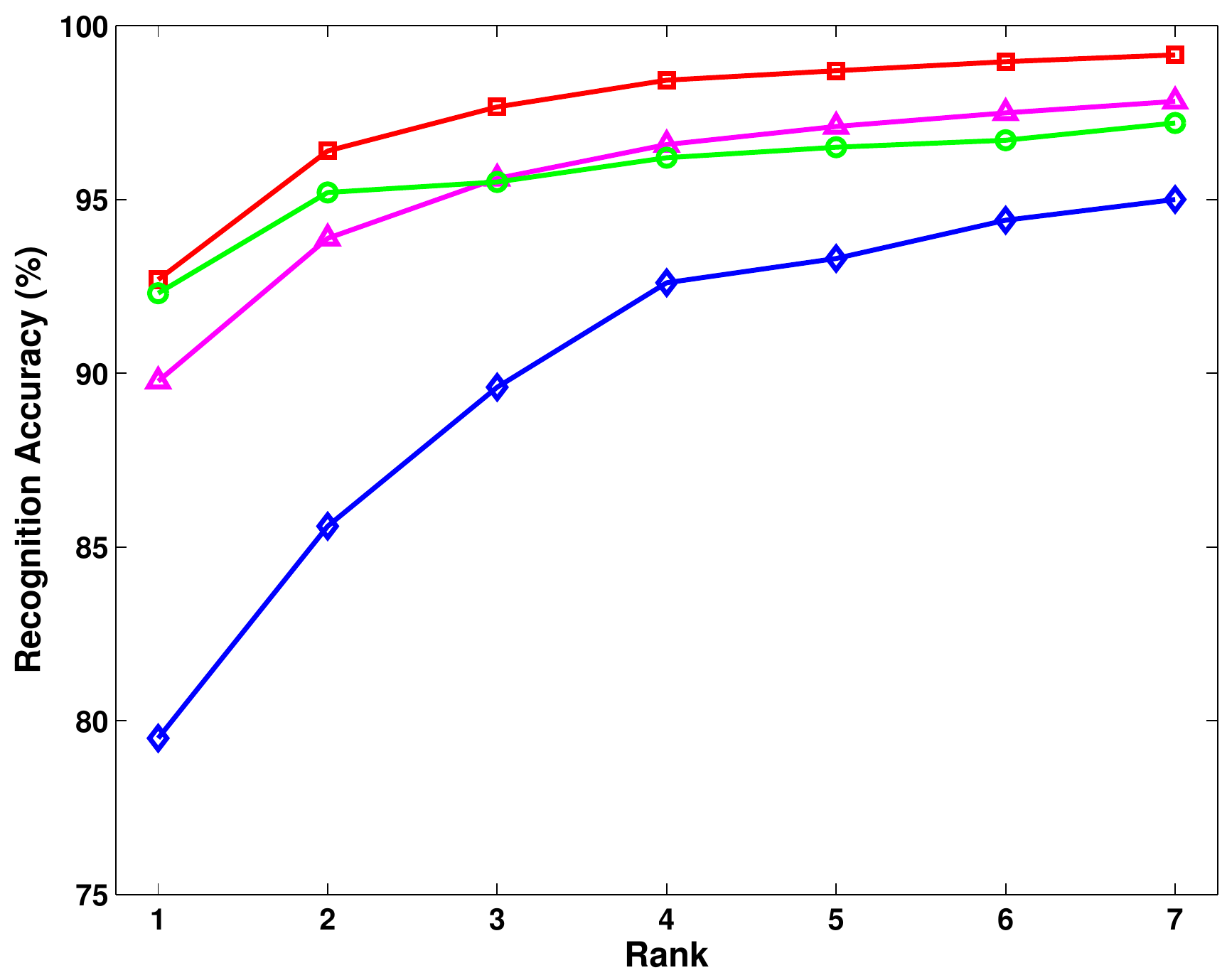}\\
      \end{minipage}
    \end{minipage}
  \caption
    {
    Performance on Sequences 1 and 2 of the ETHZ dataset (left and right panels, respectively),
    in terms of Cumulative Matching Characteristic curves.
    The proposed RSR method is compared with
    Histogram Plus Epitome (HPE)~\cite{HPE_ICPR2010},
    Symmetry-Driven Accumulation of Local Features (SDALF)~\cite{SDALF_CVPR2010}
    and Riemannian Locality Preserving Projection (RLPP)~\cite{Harandi_WACV_2012}.
    }
  \label{fig:ETHZ_results}
  \hrule
\end{figure}

\subsubsection{Person Re-identification.}
\label{sec:exp_person}

We used the modified ETHZ dataset~\cite{Schwartz_ETHZ}.
The original ETHZ dataset was captured using a moving camera~\cite{ETHZ_ICCV},
providing a range of variations in the appearance of people.
The dataset is structured into 3 sequences.
Sequence~1 contains 83 pedestrians (4,857 images),
Sequence~2 contains 35 pedestrians (1,936 images),
and
Sequence~3 contains 28 pedestrians (1,762 images).
See Fig.~\ref{fig:ETHZ_examples} for examples.

We downsampled all images to \mbox{\small $64 \times 32$} pixels.
For each subject we randomly selected 10 images for training and used the rest for testing.
Random selection of training and testing data was repeated 20 times to obtain reliable statistics.
To describe each image, the covariance descriptor was computed using the following features:
\begin{footnotesize}
\begin{equation*}
F_{x,y}
\mbox{=}
\left[~
  x,~ y,~
  R_{x,y},~   G_{x,y},~   B_{x,y},~
  R_{x,y}',~  G_{x,y}',~  B_{x,y}',~
  R_{x,y}'',~ G_{x,y}'',~ B_{x,y}''~
\right]
\end{equation*}%
\end{footnotesize}%

\noindent
where {\small $(x,y)$} is the position of a pixel,
while
{\footnotesize $R_{x,y}$}, {\footnotesize $G_{x,y}$} and {\footnotesize $B_{x,y}$}
represent the corresponding colour information.
The gradient and Laplacian for colour {\footnotesize $C$}
are represented by
{\footnotesize $C_{x,y}' \mbox{=} \left[ \left|{\partial C} \middle/ {\partial x}\right|, \left|{\partial C} \middle/ {\partial y}\right| \right]$}
and
{\footnotesize $C_{x,y}'' \mbox{=} \left[ \left|{\partial^2 C} \middle/ {\partial x^2}\right|, \left|{\partial^2 C} \middle/ {\partial y^2}\right| \right]$},
respectively.

We compared the proposed RSR method with
several techniques previously used for pedestrian detection:
Histogram Plus Epitome (HPE)~\cite{HPE_ICPR2010},
Symmetry-Driven Accumulation of Local Features (SDALF)~\cite{SDALF_CVPR2010},
and Riemannian Locality Preserving Projection (RLPP)~\cite{Harandi_WACV_2012}.
The performance of logE-SR was below HPE method and is not shown.
The results for TSC could not  be generated in a timely manner,
due to the heavy computational load of the algorithm.

Results for Sequence 1 and 2 are shown in Fig.~\ref{fig:ETHZ_results},
in terms of cumulative matching characteristic (CMC) curves.
The CMC curve represents the expectation of finding the correct match in the top {\small $n$} matches.
The proposed method obtains the highest accuracy.
For Sequence 3 (not shown),
very similar performance is obtained by SDALF, RLPP and the proposed RSR,
with HPE having the lowest performance.

\subsection{Dictionary Learning}
\label{sec:exp_dic_learning}

Here we compare the performance of the proposed Riemannian dictionary learning technique (as described in Section~\ref{sec:dic_learning}),
with the performances of dictionaries obtained by random sampling and Riemannian {\it k}-means.
We first use synthetic data to show that the proposed method obtains a lower representation error in RKHS,
followed by classification experiments on texture data.

\subsubsection{Synthetic Data.}
\label{sec:exp_dic_syn_data}

We synthesised 512 Riemannian samples from a set of 32 source points in {\small $Sym_+^5$}.
The source points can be considered as a form of ground-truth.
The synthesised samples were then used for dictionary creation
by Riemannian {\it k}-means~\cite{VIDAL:CVPR:2008} and the proposed algorithm.

To generate each source point,
an SPD matrix was created by computing
the covariance of 100 random samples of a 5 dimensional normal distribution.
The mean and variance of the distribution are different for each source point.
To synthesise each of the 512 Riemannian samples,
we uniformly selected {\small $T = 4$} source points
and combined them with random positive weights,
where the weights obeyed a normal distribution with zero mean and unit variance.

\begin{figure}[!t]
\centering
\begin{minipage}{0.55\textwidth}
  \centering
  \includegraphics[width=1\textwidth,keepaspectratio]{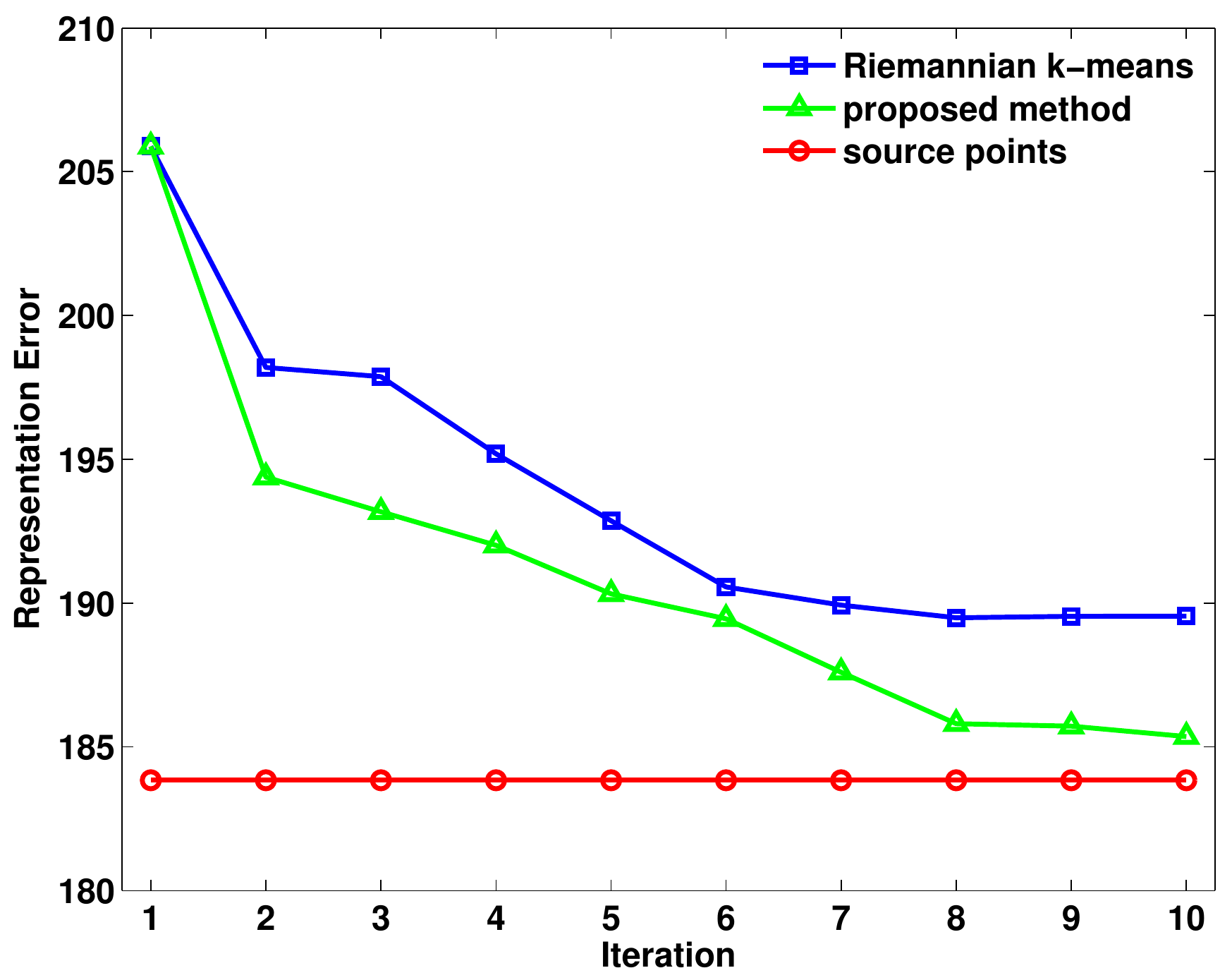}
\end{minipage}
\hfill
\begin{minipage}{0.4\textwidth}
  \caption
    {
    \small
    Representation error of learned dictionaries in RKHS, Eqn.~\eqref{eqn:dic_lrn1}, for synthetic data.
    The proposed method (Section~\ref{sec:dic_learning})
    is compared with Riemannian {\it k}-means \cite{VIDAL:CVPR:2008}.
    The source points can be interpreted as a form of ground-truth.
    }
  \label{fig:dic_learning_error_curve}

\end{minipage}
\end{figure}

The performance is measured in terms of representation error in RKHS, \ie Eqn.~\eqref{eqn:dic_lrn1}.
Fig.~\ref{fig:dic_learning_error_curve} shows the representation error as the algorithms iterate,
with the proposed algorithm obtaining a lower error than Riemannian {\it k}-means.

\subsubsection{Texture Classification.}
\label{sec:exp_dic_texture}

Here we consider a multi-class classification problem,
using 111 texture images of the Brodatz texture dataset~\cite{Brodatz_Dataset}.
From each image we randomly extracted 50 blocks of size {\small $32 \times 32$}.
To train the dictionary, 20 blocks from each image were randomly selected,
resulting in a dictionary learning problem with 2200 samples.
From the remaining blocks, 20 per image were used as probe data and 10 as gallery samples.
The process of random block creation and dictionary generation was repeated twenty times.
The average recognition accuracies over probe data are reported here.
In the same manner as in Section~\ref{sec:exp_rsr_closed_set},
we used the feature vector
\mbox
  {%
  \footnotesize%
  $F(x, y)
  \mbox{=}
  \left[
    I\left(x,y\right),
    \left| \frac{\partial I}  {\partial x}  \right|,  \left|\frac{\partial I}  {\partial y}  \right|,
    \left| \frac{\partial^2 I}{\partial x^2}\right|,  \left|\frac{\partial^2 I}{\partial y^2}\right|
  \right]$%
  }
to create the covariance,
where the first dimension is the grayscale intensity,
and the remaining dimensions capture first and second order gradients.

We used the proposed RSR approach to obtain the sparse codes,
coupled with a dictionary generated via three separate methods:
random dictionary generation,
Riemannian {\it k}-means algorithm~\cite{VIDAL:CVPR:2008},
and the proposed learning algorithm~(Section \ref{sec:dic_learning}).
The sparse codes were then classified using a nearest-neighbour classifier.

For the randomly generated dictionary case,
the classification rates are averaged over 10 runs,
with each run using a different random dictionary.
For all methods, dictionaries of size {\small $k \mbox{=} \left\{ 8, 16, 24, \cdots\hspace{-1pt}, 128 \right\}$} were trained.
The best results for each approach (\ie, the results for the dictionary size that obtained the highest recognition accuracy)
are reported in Table~\ref{tab:table_texture_dic}.
For the random dictionary, {\small $k = 64$};
for the {\it k}-means dictionary, {\small $k = 96$};
for the proposed dictionary learning algorithm, {\small $k = 24$}.
The results show that the proposed algorithm leads to a considerable gain in accuracy.

\begin{table}[!tb]
  \centering
    \begin{tabular}{ccc}
    \toprule
    ~~~~~~{\bf random }~~~~~~ &~~~~~~{\bf {\it k}-means}~~~~~~ &~~~~~~{\bf learning}~~~~~~  \\
    \toprule
    $46.09 \pm 1.5$      &$53.20 \pm 1.1$   &$60.65 \pm 0.9$ \\
    \bottomrule
    \end{tabular}

  ~

  \caption
    {
    \small
    Recognition accuracy (in \%) for the texture classification task with dictionary learning.
    In all cases the proposed RSR approach was used,
    coupled with a dictionary generated via three separate methods:
    random dictionary generation,
    Riemannian {\it k}-means~\cite{VIDAL:CVPR:2008},
    and the proposed learning algorithm~(Section \ref{sec:dic_learning}).
    }
  \label{tab:table_texture_dic}
\end{table}

\section{Main Findings and Future Directions}
\label{sec:conclusions}

With the aim of addressing sparse representation on Riemannian manifolds,
proposed to seek the solution through embedding the manifolds into RKHS,
with the aid of the recently introduced Stein kernel.
This led to a relaxed and extended version of the Lasso problem~\cite{ELAD_SR_BOOK_2010} on Riemannian manifolds.

Experiments on several classification tasks (face recognition, texture classification, person re-identification)
show that the proposed approach achieves notable improvements in discrimination accuracy,
in comparison to state-of-the-art methods such as tensor sparse coding,
Riemannian locality preserving projection,
and symmetry-driven accumulation of local features.
We conjuncture that this stems from better exploitation of Riemannian geometry,
as the Stein kernel is related to geodesic distances via a tight bound.
The proposed sparse coding method is also
considerably faster than the state-of-the-art MAXDET reformulation used by Tensor Sparse Coding~\cite{Sivalingam:ECCV:2010}.

We have furthermore proposed an algorithm for learning a Riemannian dictionary,
closely tied to the Stein kernel.
In comparison to Riemannian {\it k}-means~\cite{VIDAL:CVPR:2008},
the proposed algorithm obtains a lower representation error in RKHS
and leads to improved classification accuracies.

Future directions include using the Stein kernel for solving large margin classification problems on Riemannian manifolds.
This translates to designing a machinery that maximises a margin on SPD matrices based on Stein divergence,
which can be considered as an extension of support vector machines~\cite{Shawe-Taylor:2004:KMP} to tensor spaces.

\renewcommand{\baselinestretch}{0.97}\small\normalsize

\vspace{2ex}
\footnotesize

\noindent
{\bf Acknowledgements}.
NICTA is funded by the Australian Government as represented by the Department of
Broadband, Communications and the Digital Economy, as well as by the Australian
Research Council through the ICT Centre of Excellence program.
\vspace{-2ex}

\bibliographystyle{splncs}
\bibliography{references}

\end{document}